\documentclass[10pt,twocolumn,letterpaper]{article}

\usepackage{cvpr}
\usepackage{times}
\usepackage{epsfig}
\usepackage{graphicx}
\usepackage{amsmath}
\usepackage{amssymb}
\usepackage{caption}
\usepackage{url}

\usepackage[breaklinks=true,bookmarks=false]{hyperref}

\cvprfinalcopy 

\def\cvprPaperID{108} 

\ifcvprfinal\fi

\begin{document}

\title{WiCV 2019: The Sixth Women In Computer Vision Workshop}

 \author{
 Irene Amerini$^1$, Elena Balashova$^2$, Sayna Ebrahimi$^3$, Kathryn Leonard$^4$, Arsha Nagrani$^5$, Amaia Salvador$^6$\\\\
 $^1$University of Florence, $^2$Princeton University, $^3$UC Berkeley, \\ $^4$Occidental College, $^5$ University of Oxford, $^6$ Universitat Politecnica de Catalunya\\
 \tt\small wicv19-organizers@googlegroups.com 
}



\maketitle

\begin{abstract}
\thispagestyle{empty}
In this paper we present the Women in Computer Vision Workshop - WiCV 2019, organized in conjunction with CVPR 2019. This event is meant for increasing the visibility and inclusion of women researchers in the computer vision field. Computer vision and machine learning have made incredible progress over the past years, but the number of female researchers is still low both in academia and in industry. WiCV is organized especially for the following reason: to raise visibility of female researchers, to increase collaborations between them, and to provide mentorship to female junior researchers in the field. In this paper, we present a report of trends over the past years, along with a summary of statistics regarding presenters, attendees, and sponsorship for the current workshop.
\end{abstract}

\section{Introduction}
Tremendous research progress has been made in computer vision in the recent years over a wide range of areas, including object recognition, object detection, image understanding, video analysis, 3D reconstruction and autonomous driving. A question remains whether similar progress has been made in the inclusion of all the members of this important community. Despite the expansion of our field, the percentage of female researchers both in academia and in industry is still quite low. As a result, most female computer vision researchers feel isolated and many workspaces remain unbalanced due to the lack of inclusion.

The workshop on Women in Computer Vision is a gathering for both women and men working in computer vision targeting a broad and diverse audience of researchers, including graduate students pursuing masters or doctoral studies, undergraduates interested in research, and researchers in both industry and academia (faculty, postdocs, etc).  This workshop is envisioned as a unique opportunity to raise the profile and visibility of female computer vision researchers at all levels, seeking to reach women from different universities, with research programs and backgrounds from all around the world. 
The workshop is intended as a place of inclusion so we encourage both male and female researchers to attend the event. WiCV has three key objectives.\\
The first to maintain and increase the WiCV network the WiCV network, so that female students and professionals may share experiences and career advice. A mentoring banquet associated to the workshop provides a casual environment where those relationships can grow, providing an informal setting in which junior women can meet, exchange ideas, and form beneficial relationships with senior faculty and researchers in the field. \\ 
The second objective of the workshop is to raise visibility of women in computer vision. We invite junior and senior female researchers to give high quality research talks to present their work as keynote speakers or oral presenters. The presentations of junior researchers are of novel or ongoing scientific findings selected through peer review process. We try, as much as possible, to strive for diversity in the selected research topics and presenters' backgrounds. The workshop gives junior female researchers an opportunity to present their work in a professional and supportive setting. The keynote presentations from senior women present role-models in the shared research area with whom newcomers can have both a neutral ground for collaboration and a pathway to follow in their own career paths. The workshop also includes a panel, where the topics of inclusion and diversity can be discussed between female and male colleagues.\\
Finally, the third objective is to provide travel awards to junior female students and researchers, to simplify and encourage their participation in major vision conferences, and computer vision more broadly. These junior researchers present their work in the workshop via a poster session. In particular in this edition as well as in the last one, we have invited the submission of three page short papers on new or previously published work from female researchers, which is then presented during the poster session or oral presentation and will appear in the workshop proceedings.

\section{Workshop Program}
\label{program}

The workshop program will include 4 keynotes, 6 oral presentations, 58 poster presentations, and a panel discussion. This year, we selected our keynote speakers to balance ratios of academia versus industry, as well as junior versus senior researchers.  It is of crucial importance to diversify the set of speakers in order to provide junior researchers with different potential role models who can help them envision their own career paths. So we put special effort to diversify speakers covering different research domains and backgrounds.\\
The details of the workshop schedule are given in the following:

\begin{itemize}
  \setlength\itemsep{0.3em}
\item Introduction
\item Invited Talk 1: Raquel Urtasun (Uber ATG)
\item Oral Session 1
\begin{itemize}
\item Sara Iodice (Imperial College), \textit{Partial Person Re-identification with Alignment and Hallucination}
\item Parita Pooj (Columbia University), \textit{The Minimalist Camera}
\item Miriam Bellver (Barcelona Supercomputing Center), \textit{Budget-aware Semi-Supervised Semantic and Instance Segmentation}
\item Sadaf Gulshad (University of Amsterdam), \textit{Interpreting Adversarial Examples Using Attributes}
\end{itemize}
\item Invited Talk 2: Devi Parikh (Georgia Tech)
\item Poster Session
\item Invited Talk 3: Katie Bouman (Caltech)
\item Oral Session 2
\begin{itemize}
\item Patsorn Sangkloy (Georgia Tech), \textit{Generate Semantically Similar Images with Kernel Mean Matching}
\item Xiaodan Hu (University of Waterloo), \textit{RUNet: A Robust Architecture for Image Super-Resolution}
\end{itemize}
\item Invited Talk 4: Judy Hoffman (Facebook AI)
\item Panel Session
\item Closing Remarks
\item Mentoring Dinner 
\begin{itemize}
\item Speakers: Anima Anandkumar (Caltech),  (UC Berkeley), Cordelia Schmidt (INRIA)
\end{itemize}
\end{itemize}

\section{Workshop Statistics}

The first edition of WiCV was held at CVPR 2015. It had a great participation with a similar number of participants for the 2016 edition and an increased participation of 30\% for 2017. In 2018 two editions of WiCV were organized at CVPR and ECCV. Over the years, the number and quality of submissions to WiCV has significantly increased, we foresee a large number of participants for the next edition of the workshop as well. Due to the large number of submissions this year, we were encouraged to collect the top quality submissions into workshop proceedings, following the example of last year's workshops \cite{Akata18,Demir18}. We believe that this will increase the visibility of female researchers even more by providing our oral and poster presenters with a permanent publication record of their participation. 
This year, the workshop is a half day gathering. Senior and junior researchers have been invited to present their work, and poster presentations are included as already described in the previous Section \ref{program}.\\

In this edition of WiCV, the organizers are from geographically distributed institutions in four different timezones, with a maximum of nine hours difference. The organizers are coming from academia, and each was born in a different country. Moreover, their research areas of interest are also diverse, including computer vision, computer graphics, machine learning, computational geometry, multimedia forensics, and robotics, which brought many perspectives to the organizing committee.

\begin{figure}[h]
    \centering
    \includegraphics[width=1\linewidth]{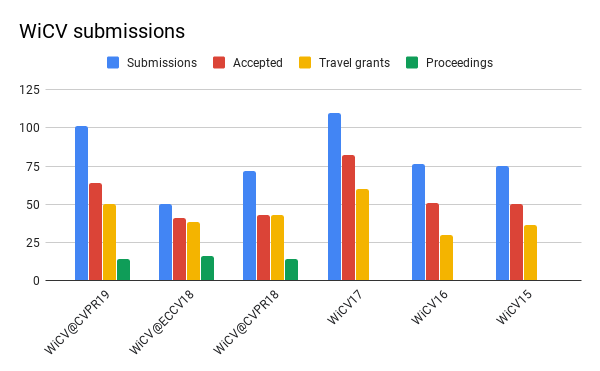}
    \captionof{figure}{\textbf{WiCV Submissions.} The number of submissions over the past years of WiCV.}
    \label{fig:sub}
\end{figure}

This year we had 101 high quality submissions from a wide range of topics and institutions. The most popular topics were object recognition, detection and categorization, applications of computer vision, deep learning and convolutional neural networks.
There were also very interesting research applications ranging from modeling 3D buildings from satellite imagery, to facial expression synthesis, to adversarial neural networks. Over all submissions, around 6\% were selected to be presented as oral talks, 14\% were selected to be included in the proceedings, and 64\% were selected to be presented as posters. The comparison with previous years is presented in Figure~\ref{fig:sub}. We had a broad program committee of 74 reviewers to evaluate and to improve the papers with fruitful comments.

This year we were able to continue the positive tradition of WiCV at CVPR and ECCV 2018 in providing travel grants for all the authors of accepted submissions who requested funding. 

\begin{figure}
    \centering
    \includegraphics[width=1\linewidth]{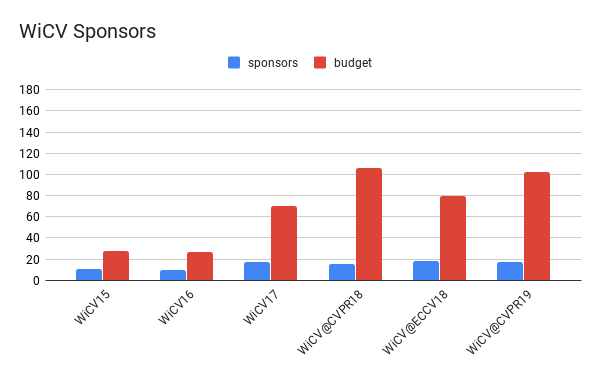}
    \captionof{figure}{\textbf{WiCV Sponsors.} The number of sponsors and the amount of sponsorships for WiCV. The amount is expressed in US dollar (USD).}
    \label{fig:spo}
\end{figure}

The total amount of sponsorship this year is \$102.500 USD with 17 sponsors, reaching a very good target. In Figure~\ref{fig:spo} you can find the details respect to the past years. The majority of this sponsorship will be spent on travel stipends, and the rest will be used to cover the expenses of the mentorship banquet, workshop bags and promotional materials, and some travel support for organizers. 

\section{Conclusions}
WiCV at CVPR 2019 has continued to be a valuable opportunity for presenters, participants and organizers in its contributions to build a more connected community and to overcome the ongoing gender imbalance. Given the high number of submitted papers and attendees we foresee that the workshop will continue the marked path of previous years in increasing visibility, providing support, and building a strong community for all the female researchers in academy and in industry.

\section{Acknowledgments}
First of all we would like to thank our sponsors. Our banquet sponsor Google allowed us us to organize a wonderful mentorship dinner on the Long Beach coast. We are also grateful to our other Platinum sponsors: Amazon, Facebook, IBM Research and Microsoft. We would also like to acknowledge our Gold sponsors: Apple, Toyota Research Institute, Uber; Silver Sponsors: Disney Research, Intel AI, Lyft and Waymo as well as our Bronze sponsors: Adobe, Argo, Magic Leap, Mitsubishi Electric and Siemens. We would also would like to thank Occidental College as our fiscal sponsor, which donated employee knowledge and time to process our sponsorships and travel awards. We would also like to thank and acknowledge Ilke Demir for the useful tips at the beginning of our workshop planning and Dena Bazazian for the incredible help throughout all the organization. Without the information flow and support from the previous WiCV organizers, this WiCV would not have been possible. We would like to also acknowledge CVPR 2019 Workshop Chairs, Sanja Fidler and Andrea Vedaldi, the Publications Chairs William Brendel and Mohamed R. Amer for answering all concerns timely, as well as Eric Mortensen for answering all IEEE publishing questions. Finally, we would like to acknowledge the time and efforts of our program committee, authors, submitters, and our prospective participants for being part of WiCV network community.

\section{Contact}
\noindent \textbf{Website}: \url{http://wicvworkshop.github.io}\\
\textbf{E-mail}: wicv19-organizers@googlegroups.com\\
\textbf{Facebook}: \url{https://www.facebook.com/WomenInComputerVision/}\\
\textbf{Twitter}: \url{https://twitter.com/wicvworkshop}\\

{\small
\bibliographystyle{ieee}
\bibliography{egbib}
}

\end{document}